\documentclass[letterpaper, 10pt, conference]{ieeeconf}

\IEEEoverridecommandlockouts
\usepackage{graphics}

\usepackage{amsmath}
\usepackage{amssymb}
\usepackage[colorlinks, linkcolor=blue, citecolor=blue]{hyperref}
\usepackage{cite}

\usepackage{algorithm}
\usepackage{url}
\usepackage{algpseudocode}
\usepackage{epsfig}
\usepackage{graphicx}
\usepackage{caption}
\usepackage{subfigure}
\usepackage{mathrsfs}
\usepackage{array}
\usepackage{multirow}
\usepackage{multicol}
\usepackage{booktabs}
\usepackage{makecell}
\usepackage{xcolor}
\usepackage{amsfonts}
\usepackage{gensymb}
\usepackage{threeparttable}

\title{\LARGE \bf
Learning Observation-Based Certifiable Safe Policy for \\ Decentralized Multi-Robot Navigation
}

\author{Yuxiang Cui, Longzhong Lin, Xiaolong Huang, Dongkun Zhang, Yue Wang, Rong Xiong
\thanks{All authors are with the State Key Laboratory of Industrial Control and Technology, Zhejiang University, Hangzhou, P.R. China. Yue Wang is the corresponding author.{\tt\small wangyue@iipc.zju.edu.cn}.}
}

\begin{document}

\maketitle
\thispagestyle{empty}
\pagestyle{empty}

\begin{abstract}
Safety is of great importance in multi-robot navigation problems. 
In this paper, we propose a control barrier function (CBF) based optimizer that ensures robot safety with both high probability and flexibility, using only sensor measurement.
The optimizer takes action commands from the policy network as initial values and then provides refinement to drive the potentially dangerous ones back into safe regions.
With the help of a deep transition model that predicts the evolution of surrounding dynamics and the consequences of different actions, the CBF module can guide the optimization in a reasonable time horizon.
We also present a novel joint training framework that improves the cooperation between the Reinforcement Learning (RL) based policy and the CBF-based optimizer both in training and inference procedures by utilizing reward feedback from the CBF module.
We observe that the policy using our method can achieve a higher success rate while maintaining the safety of multiple robots in significantly fewer episodes compared with other methods.
Experiments are conducted in multiple scenarios both in simulation and the real world, the results demonstrate the effectiveness of our method in maintaining the safety of multi-robot navigation. 
Code is available at \url{https://github.com/YuxiangCui/MARL-OCBF}.

\end{abstract}

\section{Introduction}
Reinforcement learning (RL) has shown great success in the domain of autonomous navigation.
As for decentralized multi-robot navigation tasks, not only should the policy pay attention to the safety between the robots and the environment, but also it should take the safety of robot interactions into consideration.
RL-based methods set specific reward functions to encourage each agent to make decisions considering the evolution of the surrounding dynamics while approaching its targets. 
However, the safety of these methods can hardly be guaranteed in deployment due to the prime pursuit for long-term return. 

According to the network input type, RL-based multi-robot navigation algorithms can be roughly divided into state-based methods and observation-based methods. 
In state-based methods, the world is simplified to discrete points with specific states like position and velocity \cite{everett2018motion}\cite{riviere2020glas}.
While the high dependency on accurate detection constrains their performance in real-world applications. 
However, observation-based methods can encode an arbitrary number of obstacles using only raw sensor data, leveraging the permutation invariant nature of multi-robot interactions\cite{fan2018fully}\cite{jin2020mapless}\cite{cui2020learning}.
These methods have achieved quite stunning performance in multi-robot navigation tasks with a high success rate, while collisions are still inevitable.

\begin{figure}[t]
\centering
\includegraphics[scale=0.35]{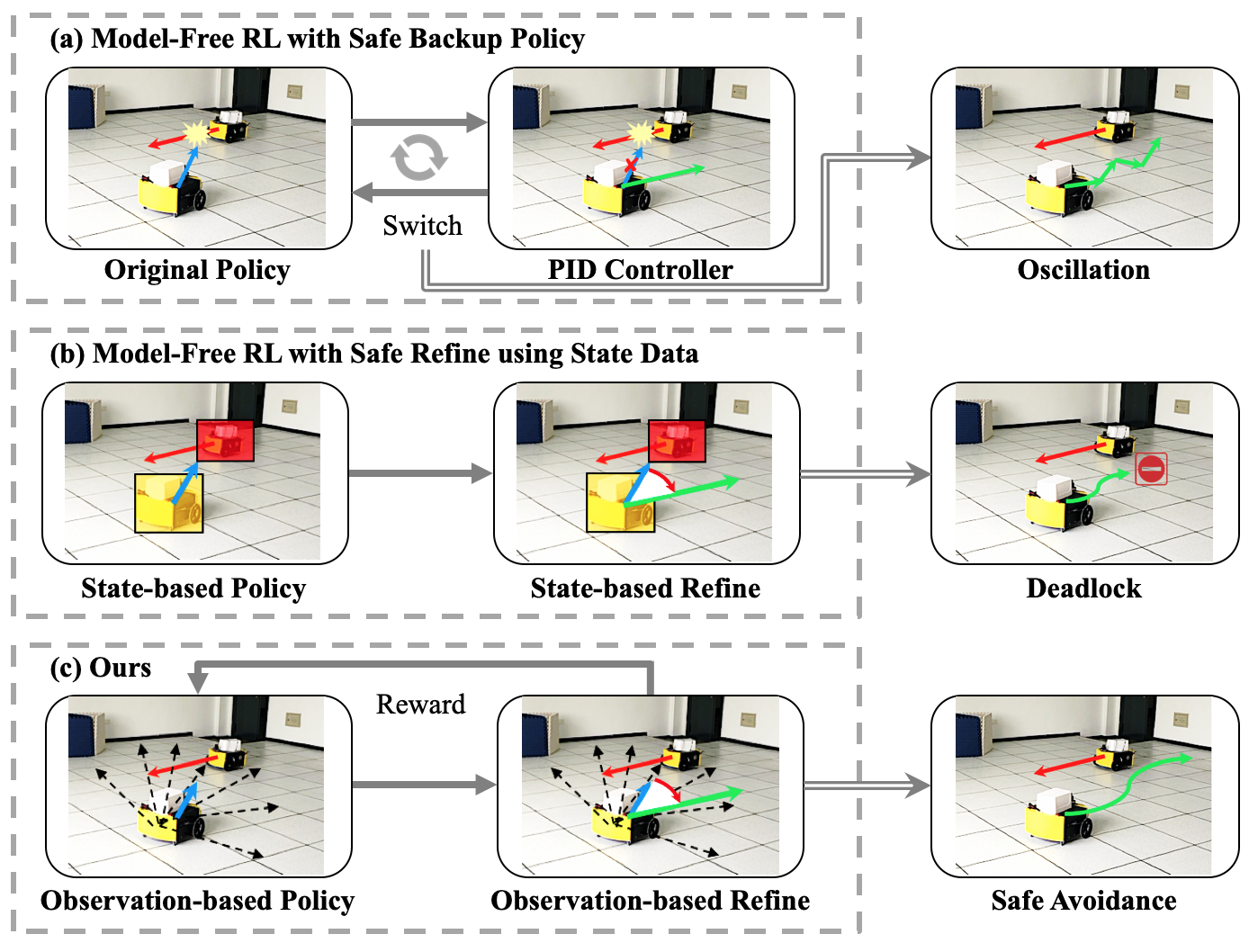}
\caption{\small{Illustration comparing (a)model-free RL with safe backup policy, (b)model-free RL with safe refine using state data, and (c)our observation-based safe policy. Note that our method utilizes only observation data to ensure safety and uses the safe module to provide both refinement and reward feedback for policy improvement, instead of frequent switching to backup policy, leading to better cooperation and safe performance.}}
\label{First_img}
\vspace{-0.7cm}
\end{figure}

On the other side, safety is also important for multi-robot navigation problems. 
Methods like using PID controller as backup policy\cite{fan2018fully} emerge these years, but the frequent switching policies usually lead to problems like oscillations. 
Control Barrier Functions (CBF)\cite{ames2019control} are introduced to multi-robot systems recently\cite{qin2021learning}\cite{cheng2020safe}.
They guarantee safety with the property of forward invariance, meaning that the system keeps staying in the safe region.
However, these methods usually solve the problems in state space, where the accurate positions or velocities are precisely known, which could be quite unrealistic in real-world applications.
And the mismatch between the policy and refine modules in an open-loop framework may result in deadlock.

In this paper, we develop a novel joint training framework that utilizes the complementary strength of both observation-based CBF and RL for decentralized safe multi-robot navigation.
With the help of a world transition model, the CBF module can estimate the state of the robots under the influence of the policy in multi-step horizon using only sensor measurement, which guides the optimizer to actions with minimal violation of safety constraints.
In this way, we can bridge the gap between CBF and the observation representation efficiently.
To enhance cooperation, we use CBF to provide both refinement and reward feedback for the policy, leading to safety certificates in both training and inference procedures with substantially improved efficiency.
Under this connection, the policy and the CBF-based optimizer can cooperate on the basis of a unified cognition.
Taking advantage of the RL, our method can also avoid local minima like multi-robot deadlock, which is common in decentralized methods. 

To the best of our knowledge, this is the first work that utilizes a joint training framework of observation-based CBF with RL for safety certificates in multi-robot navigation. 
Particularly, this paper presents the following contributions:

\begin{itemize}
	\item We propose a joint training framework for RL-based multi-robot navigation policy with CBF-based optimizer, where the optimizer provides both action refinement for safety certificates and reward feedback for the improvement of policy cooperation. 
	
	\item We design an observation-based CBF that estimates the violation of safety constraints using only robot self-state and sensing data, providing gradients for policy refinement with the help of a world transition model.
	
	\item We train the policy in multiple simple simulation environments and directly evaluate it in random scenarios in both simulations and on real robots. The results show promising safe performance in real-time both in training and inference procedures.
\end{itemize}

\section{Related Works}

\subsection{Reinforcement Learning for Multi-Robot Navigation}

RL-based multi-robot navigation methods learn from interactions. 
Decentralized policy networks take local information as input and outputs the control commands.
State-based RL methods\cite{chen2017decentralized} directly use other agents' states or surrounding obstacles' states like positions and velocities, as local information.
Although with the help of network architectures like attentive pooling\cite{chen2019crowd} and LSTM\cite{everett2018motion}, these methods can deal with an arbitrary number of state inputs, the high dependence on robust detection modules still constraints their deployment on real robots. 
While observation-based methods take sensor measurements like laser scans\cite{chen2017decentralized}\cite{jin2020mapless} as network inputs to leverage the permutation invariant nature of collision avoidance. 
Our previous work also uses laser scans in stacked obstacle map representation for social navigation and further extended this method into the task of multi-robot target following using the artificial potential field\cite{cui2021socially}.
These methods can achieve high success rates in a variety of multi-robot tasks, but none of them have a formal guarantee for safety. 
The reward settings can not serve as hard constraints that ensure the safety of robots.

\subsection{RL-based multi-robot Navigation with Safety Concerns}

These years, researchers start to pay attention to the safe deployment of RL-based navigation algorithms. 
Some of them use backup policies like traditional collision avoidance algorithms\cite{fan2018fully} or model predictive shielding algorithms\cite{zhang2019mamps} for safety-critical scenarios.
Others try to separate the pipeline into two parts, RL-based subgoal recommendation and model predictive control with safety certificates\cite{brito2021go}. 
These methods utilize the complementary strength of both RL and traditional methods with safety certificates.
However, in order to make these traditional modules suitable, they usually make strong assumptions on the world dynamics or multi-robot communications\cite{cheng2020safe}.

\subsection{Autonomous Navigation using Control Barrier Functions}

Plenty of works have proved the feasibility of CBF in multi-robot navigation tasks of state-based settings\cite{wang2017safety}\cite{chen2017obstacle}\cite{riviere2020glas}\cite{wang2016safety}.
But to build an analytical expression, these methods usually set strong assumptions on the environment dynamics, like accurately known robot dynamics in single-agent problems\cite{ames2014control} and perfect global communication in multi-robot problems\cite{wang2017safety}. 
Therefore, these algorithms are usually used in specifically structured state-based problems or single-agent scenarios\cite{boffi2020learning}\cite{cheng2019end}.
Recently, researchers try to use deep learning methods to obtain CBFs from data in complex settings.
For example, CBF in observation space\cite{srinivasan2020synthesis}, CBF for models with uncertainties\cite{cheng2020safe}\cite{taylor2020learning} and CBF underlying in expert demonstrations\cite{robey2020learning}\cite{saveriano2019learning}
However, these methods either serve as state classifiers for safety concerns that lack interpretation or need a large amount of expert data.

Not only that, in these works, the policy is not considered in the process of CBF construction.
CBF works as an auxiliary module that provides refinement to the predefined nominal controllers, the stability of the system might not be kept with safety concurrently.
There are also works that jointly learn the policy with the barrier functions cooperating with it\cite{qin2021learning}\cite{jin2020neural}.
While these methods adopt the supervised learning paradigm that depend on guidance from expert demonstrations or Lyapunov-like functions, which also need failure experiences.
RL-based methods try to ensure safety even in the learning process, but the high dependence on a fast convergent transition model and compensating policy limits its scalability to complex systems with multiple robots\cite{cheng2019end}.
These works inspired us to build the framework in this paper that uses CBF for both safe certificate and policy enhancement in the observation space.

\begin{figure*}[ht]
\centering
\includegraphics[scale=0.32]{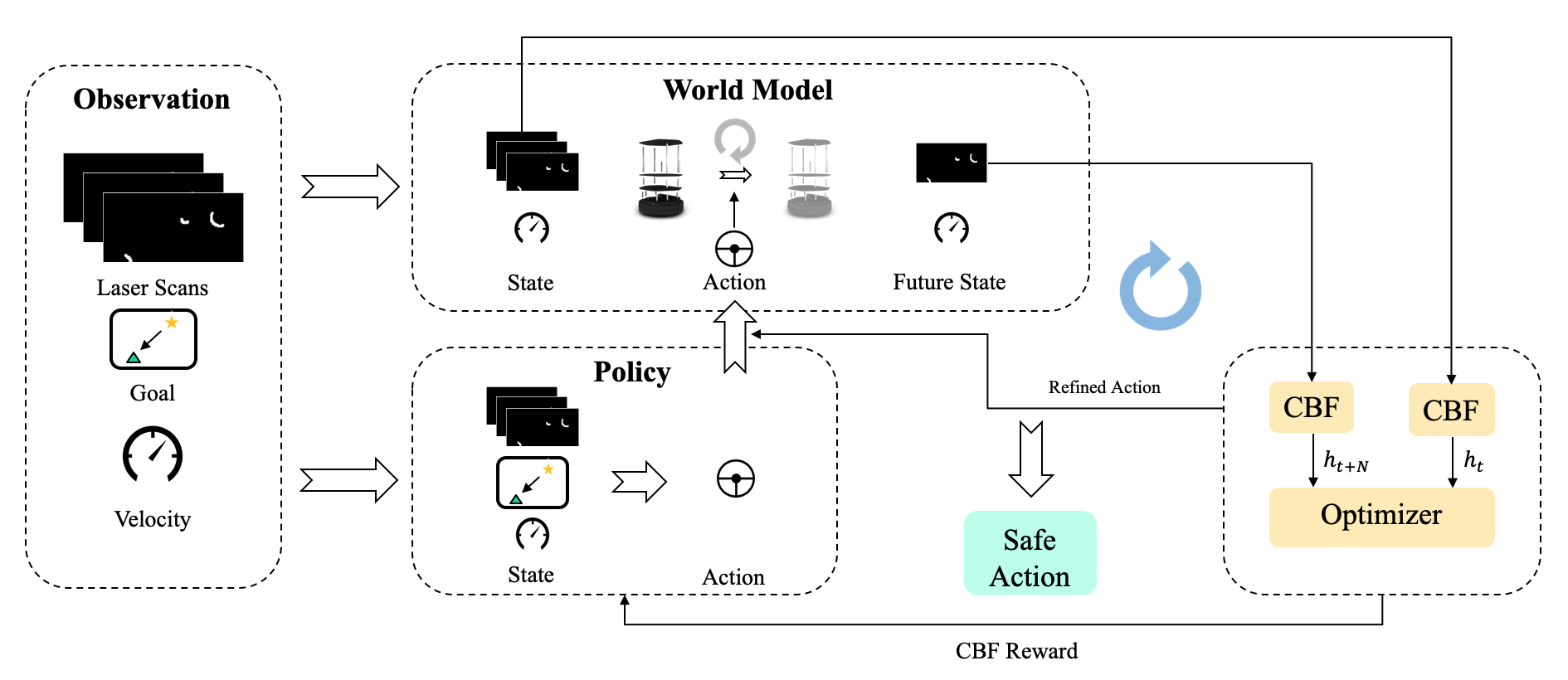}
\caption{\small{Framework of joint training. The observation space includes robots' laser scans in stacked obstacle map representation, relative goal positions, and velocities. The policy network directly maps the observations to initial actions for further refinement. The world model takes history obstacle maps and velocity as input and predicts future states iteratively in a reasonable time horizon. And then the CBF-based optimizer figures out the differences between current states and future states and then conducts refinement until the action with minimally safety violations is obtained.}}
\label{Framework}
\vspace{-0.7cm}
\end{figure*}

\section{Decentralized Observation-based CBF using World Transition Model}
\subsection{Decentralized Observation-based CBF}

In this part, we aim to design an observation-based control barrier function that estimates the safety status using only robots' sensor data and obtain the derivative condition under the influence of the current action and policy with the help of a deep world transition model. 

Instead of a centralized control barrier function that easily leads to exponential blow-up in multi-robot tasks, we utilize a decentralized control barrier function, which is also proven to be able to guarantee the safety of multi-robot systems\cite{qin2021learning}.
The decentralized control barrier function can enforce each robot to stay in its safe region, leading to a collision-free multi-robot system.

As for each robot in the multi-robot system, the evolution of its state can be defined as:
\begin{equation}
s^{(t+1)} = f(s^{(t)}, a^{(t)}),
\end{equation}
where $ s \in \mathcal{S} \subset \mathbb{R}^n$ stands for the state of the robot and $ a \in \mathcal{A} \subset \mathbb{R}^m$ stands for the action taken by the robot.

A Control Barrier Function (CBF) $h$ guarantees the safety with forward invariance of safe state set, which satisfies:
\begin{equation}
\label{eq:cbf}
\begin{aligned}
& \forall s \in \mathcal{S}_{0} \subset \mathcal{S}, h(s) \ge 0, \\
& \forall s \in \mathcal{S}_{s} \subset \mathcal{S}, h(s) \ge 0, \\
& \forall s \in \mathcal{S}_{d} \subset \mathcal{S}, h(s) < 0, \\
& \bigtriangledown_{s} h \cdot f(s, a) + \alpha(h) \ge 0, \\
\end{aligned}
\end{equation}
where $\mathcal{S}_{0}, \mathcal{S}_{s}, \mathcal{S}_{d}$ define initial state set, safe state set, and dangerous state set, respectively. Here $\bigtriangledown_{s} h \cdot f(s, a)$ is the expanded form of time derivative $\dot{h}$, and we call the last inequality in (\ref{eq:cbf}) derivative condition of CBF.

In the navigation task, every robot estimates its own state using observation from on-board sensors.
Therefore, in our task, the state of the robot is composed of the laser readings $s_l$, relative goal position $s_g$, and robot’s own velocity $s_v$, which can be directly acquired through sensors mounted on the robot. 
Here, we assume that the safety of the robot is determined by $s_l$ and $s_v$, which contain the information of collisions.

In previous work\cite{cui2020learning}, we find that the obstacle map representation of laser scans could lead to better performance of both the policy network and the world transition model.
Therefore, to cooperate CBF with them, we evaluate the safety status using $s_v$ and $s_l$ in the costmap representation.
We define a costmap-based control barrier function that estimates the pixel-wise safety in the robot coordinate system. The safety status between the robot and the grid at $(i,j) \in H \times W$ in the costmap ($s_{l,ij}$ means whether there is obstacle) is defined as:
\begin{equation}
h_{ij}(s)=
\begin{cases}
r_{ij} - v_{ij} \Delta{t} - r_{min} & s_{l,ij}=1,\\
r_{max}&s_{l,ij}=0.
\end{cases}
\end{equation}
where $r_{ij} = ||d_{ij}||_2$ stands for the distance between the robot and the obstacle at $d_{ij} = (x,y)$, $v_{ij} =  \frac{s_v^T d_{ij}}{r_{ij}}$ stands for the robot velocity projected on the direction to the obstacle at $d_{ij}$, 
$r_{min}$ stands for the minimum safe distance, $r_{max}$ stands for the maximum considered distance, $ H $ and $ W $ are respectively the height and width of the obstacle map.

\begin{figure}[t]
\centering
\includegraphics[scale=0.34]{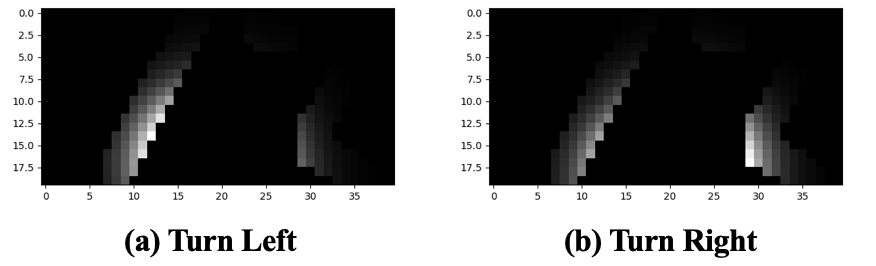}
\caption{\small{Costmap-based CBF visualization. Here we present the visualization of the CBF value distribution in the same obstacle map with two different velocity states. The brighter the grids, the more dangerous it is for the current robot.}}
\label{CBF}
\vspace{-0.2cm}
\end{figure}

This function estimates safety by checking the distance between the robot and the grids occupied with obstacles in the obstacle map, and figuring out whether the robot will collide with them in $\Delta{t}$ while the robot keeps the current velocity.
The visualization of the pixel-wise CBF is shown in Fig. \ref{CBF}. We can see that the safe estimation distribution varies with the robot's own velocity.
Larger $\Delta{t}$ and $r_{min}$ means more conservative safety standards.
Intuitively, if the robot is going to run into an obstacle at $d_{ij}$, then $h_{ij} < 0$, implying that the safety between them is about to be broken.
And as for the grids not occupied with obstacles or grids out of the safe consideration range, $h_{ij} \ge 0$.

\subsection{Derivative Conditions from World Transition Model}

\begin{figure}[t]
\centering
\includegraphics[scale=0.32]{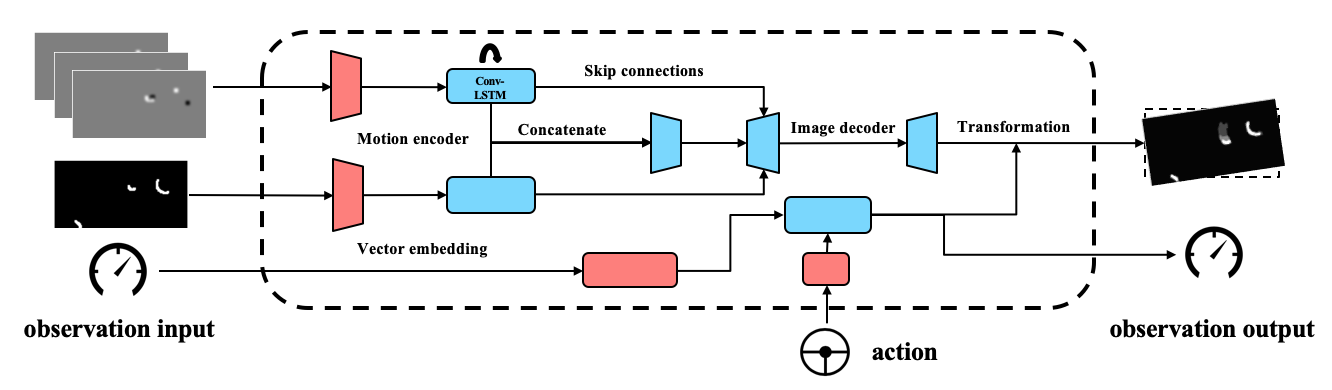}
\caption{\small{Architecture of the world transition model. The input includes laser scans in stacked obstacle map representation and the current velocity of the robot and the outputs are the predicted obstacle map and velocity. 
Specifically, the sequential obstacle maps are further processed into motion and content, in which motion is composed of differences between adjacent frames and content is represented by the last frame.}}
\label{world model}
\vspace{-0.7cm}
\end{figure}

It is insufficient to ensure the forward invariant safety of the system only through checking the current safety status using the function $h$.
Therefore, we should also check the derivative condition of CBF using the world model that predicts future observations.

As for the multi-robot problems, both the motion of dynamic obstacles in the field of view and the ego-motion of the robot itself are involved.
To predict the future observations, we build a deep world transition model $\mathcal{M}$ with the similar network structure proposed in our previous work\cite{cui2020learning} as shown in Fig. \ref{world model}.
Specifically, laser scans are processed into a stacked obstacle map representation disentangled from the robot's self-motion before input.
By further decoupling the environment dynamics from the static content, the world model can encode the sequential motion information and predict the future observations in the robot coordinate system.
Affine transformations are also used to deal with future self-motion.
As a result, the world model predicts the future obstacle map and velocity under the influence of the action chosen by the policy.

To evaluate the derivative condition of $ h_{ij} $, we define a pixel-wise safety check with the help of the world transition model.
Note that the predicted obstacle map would go through affine transformations induced by self-motion of the robot, the grids on the predicted map do not match exactly those at the same positions on the current obstacle map.
Therefore, an inverse affine transformation is needed for building pixel-wise CBF  derivative conditions in which the grids between the current and the predicted obstacle map corresponding to the same obstacles match correctly.

Then, the time derivative of $h_{ij}$ can be approximated by:
\begin{equation}
\dot{h}_{ij} \approx h^{(t+\Delta{t})}_{ij} - h^{(t)}_{ij} = \{T^{-1}[h(\mathcal{M}(s^{(t)}, a^{(t)}))]\}_{ij} - h^{(t)}_{ij},
\end{equation}
where $T^{-1}$ refers to the inverse affine transformation.
This function of $ h_{ij} $ defines a local pixel-wise safety status difference. 
As for the safety of the robot, all obstacles inside the range of the maximum considered distance should be concerned.
And the safety of the robot can be guaranteed when no collisions are detected in a time horizon $\Delta{t}$.

Actually, the world transition model can make multi-step predictions for safety status estimating with a series of actions provided by the policy, but to keep a real-time control, we only use a one-step look-ahead. 
While in comparison, model-free methods only estimate the current state.

\begin{algorithm}[tb]
\caption{Model-based RL-CBF}
\label{algorithm_cbf_rl}
\begin{algorithmic}[1]
\State Initialize policy $\pi_{\phi}$, predictive model $p_{\theta}$ and dataset $\mathcal{D}$;
\For{N epochs}
\For{K steps}
\If{total step $<$ E}
\State Take a random action $a^{(t)}$;
\Else
\State Sample $a^{(t)}$ from $\pi_{\phi}$ according to state $s^{(t)}$;
\EndIf
\If{epoch $<$ M}
\State Refine $a^{(t)}$ to $a^{(t)}_{*}$ by solving CBF-QP with static assumption, and get CBF reward $r_c^{(t)}$;
\Else
\State Refine $a^{(t)}$ to $a^{(t)}_{*}$ by solving CBF-QP with $p_{\theta}$, and get CBF reward $r_c^{(t)}$;
\EndIf
\State Apply $a^{(t)}_{*}$ and get next observation $s^{(t+1)}$;
\State Compute goal reward $r_g^{(t)}$ and total reward $r^{(t)} = r_g^{(t)} + r_c^{(t)}$;
\State Add data $(s^{(t)}, a^{(t)}, a^{(t)}_{*}, s^{(t+1)}, r^{(t)})$ to $\mathcal{D}$;
\State Update $\pi_{\phi}$ and $p_{\theta}$ on $\mathcal{D}$;
\EndFor
\EndFor

\end{algorithmic}
\end{algorithm}

\begin{figure}[tb]
\centering
\subfigure{
\centering
\includegraphics[scale=0.18]{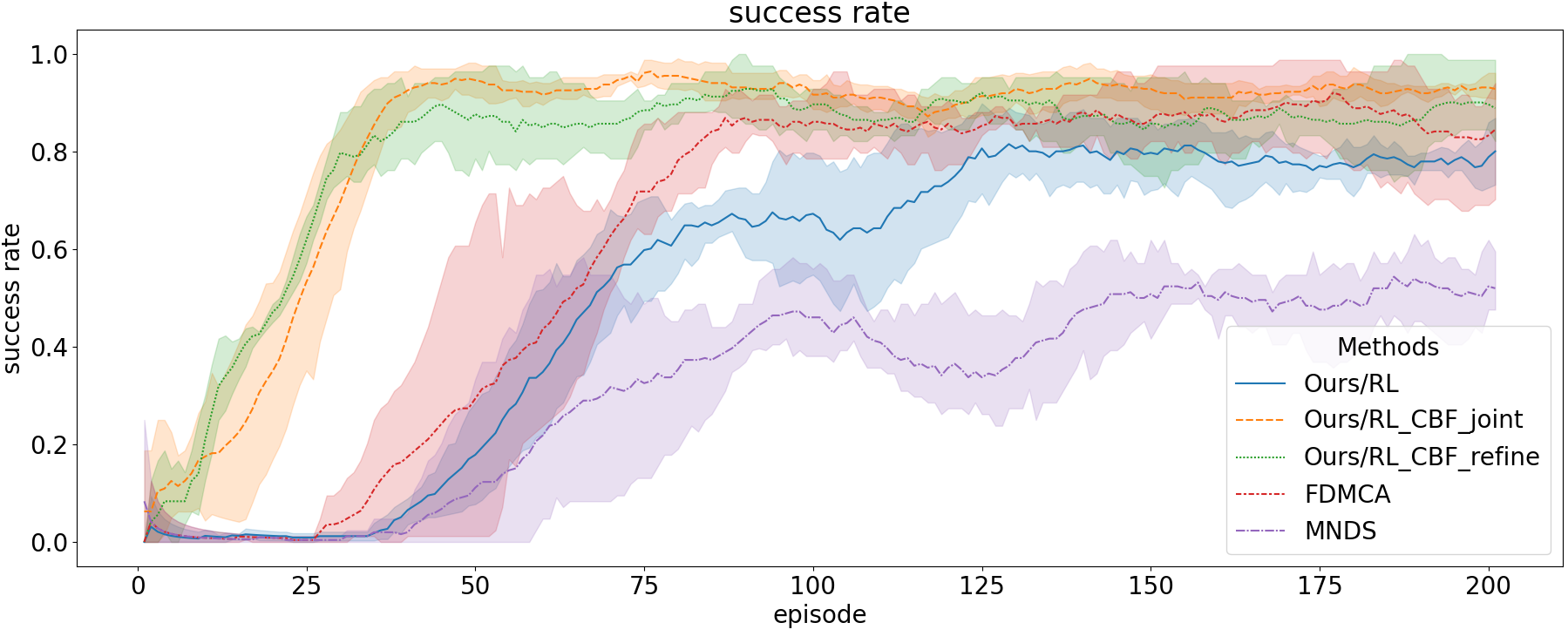}
}
\subfigure{
\centering
\includegraphics[scale=0.18]{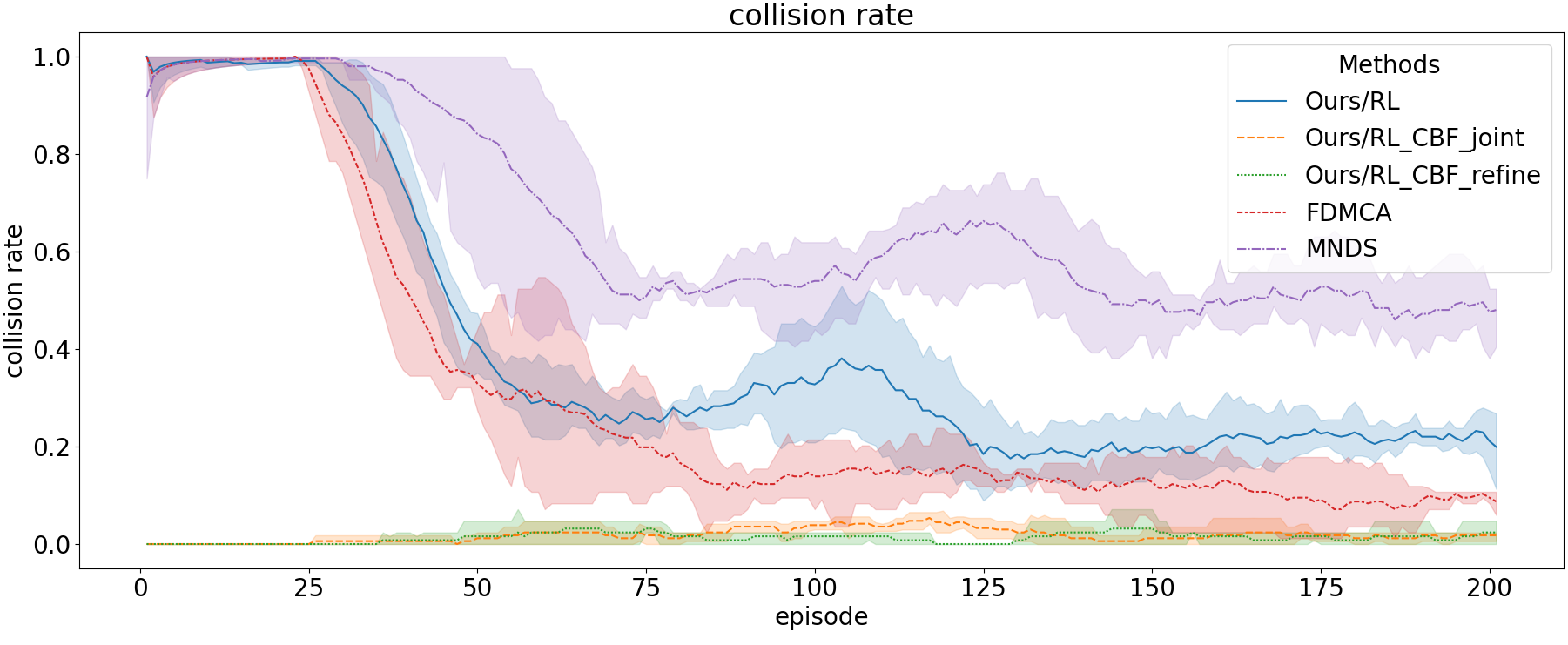}
}
\subfigure{
\centering
\includegraphics[scale=0.18]{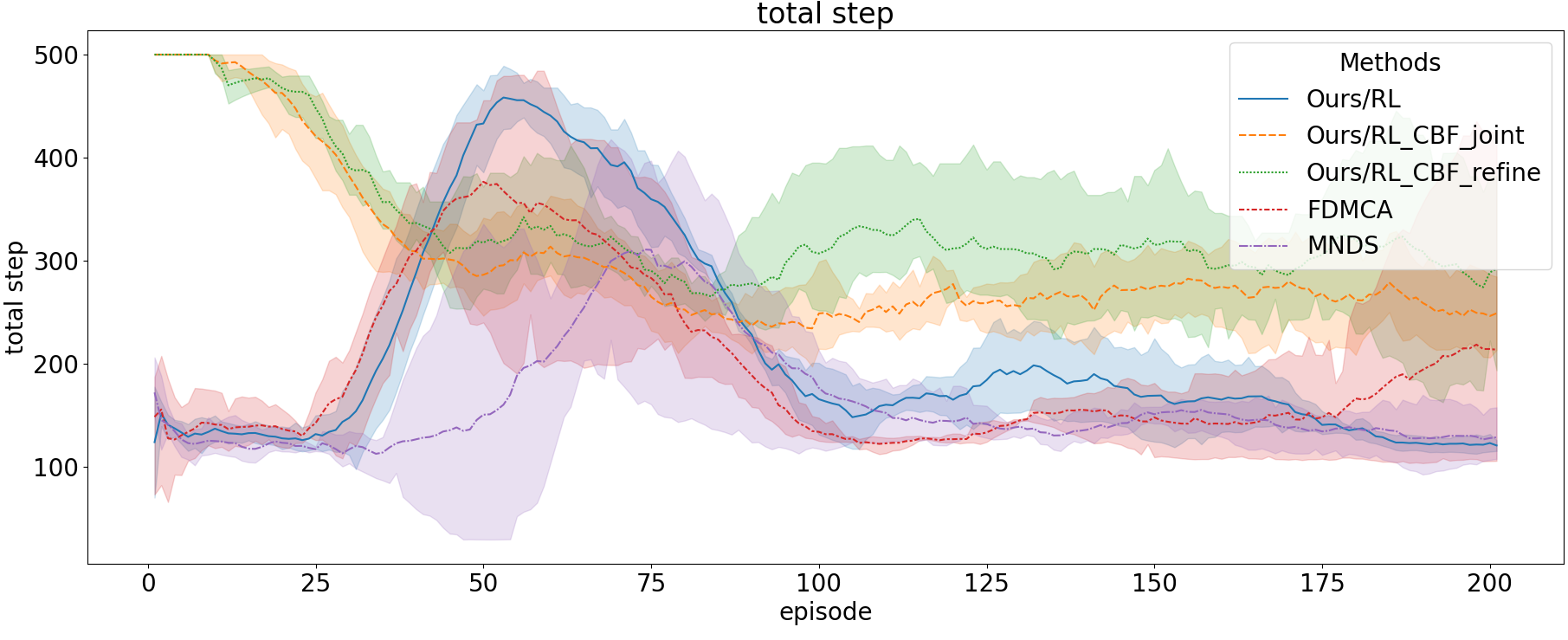}
}
\centering
\caption{\small{Comparison between our methods and the baselines on success rate, collision rate and total steps. We refer to our full version method as 'Ours/RL\_CBF\_joint' , the version without using CBF reward feedback as 'Ours/RL\_CBF\_refine' and the original method with individual RL policy as 'Ours/RL'.
The solid curves and shaded regions depict the mean and the standard deviation of average reward among three trials. We can observe that our methods can achieve a higher success rate while maintaining safety in significantly fewer episodes compared with other methods.
}}
\label{Training_Curves}
\end{figure}

\section{Joint Training Framework of RL and CBF}
\subsection{Policy Refinement}
To ensure the safety of a multi-robot navigation policy $ \pi_{\phi}(s) $ with RL, we utilize the CBF for policy refinement both in the training and inference procedure, leading to the performance of safe exploration and exploitation in RL.

Using the routine paradigm of solving a CBF constrained problem, our task can be formulated as: 
\begin{equation}
\begin{aligned}
&\min_{a} \quad  ||a-\pi_{\phi}(s)||_2^2 \\
&\begin{array}{ll}
	s.t. & h^{(t+\Delta{t})}_{ij} - h^{(t)}_{ij} \ge - \alpha h^{(t)}_{ij} - \epsilon \ , \ (i,j) \in H \times W \\
	& a_{min} \le a \le a_{max} \\
\end{array}
\end{aligned}
\end{equation}
where we should determine the actions that satisfy the constraints on both safety and robot dynamics, with a relaxation variable $ \epsilon $ that ensures the solvability.

To solve this problem with non-analytical constraints, we use the Augmented Lagrangian method and transform the problem formulation into an unconstrained optimization problem:
\begin{equation}
\begin{aligned} \label{P}
& \min_{a} \mathcal{L}_{\sigma}(a, \pi_{\phi}(s), \lambda_{ij}, \lambda_{1}, \lambda_{2}) \\
& = \min_{a} ||a-\pi_{\phi}(s)||_2^2 \\
& -(\sum_{ij} \lambda_{ij} \mathcal{C}_{CBF} 
+ \lambda_{1}\gamma||\mathcal{C}_{dyn \_ up}||_2  +\lambda_{2}\gamma||\mathcal{C}_{dyn \_ low}||_2) \\ 
& +\frac{\sigma}{2}(\sum_{ij}(\mathcal{C}_{CBF}^2 
+ (\gamma||\mathcal{C}_{dyn \_ up}||_2)^2 
+ (\gamma||\mathcal{C}_{dyn \_ low}||_2)^2) \\
\end{aligned}
\end{equation}
in which $\mathcal{C}_{CBF}$, $\mathcal{C}_{dyn \_ up}$ and  $\mathcal{C}_{dyn \_ low}$ are the CBF constraints and robot dynamic constraints defined by :
\begin{equation}
\begin{aligned}
& \mathcal{C}_{CBF} =  \min\{\Delta{h}_{ij} + \alpha h_{ij} , 0\} \\
& \mathcal{C}_{dyn \_ up} =  \min\{a_{max}-a , 0\} \\
& \mathcal{C}_{dyn \_ low} = \min\{a-a_{min} , 0\} \\
\end{aligned}
\end{equation}

where $\gamma$ is the normalization factor for robot dynamic constraint and $a$ stands for the refined action.

The parameters $\lambda, \sigma$ update periodically following:
\begin{equation}
\begin{aligned}
& \lambda_{ij} \leftarrow \lambda_{ij} - \sigma \mathcal{C}_{CBF} \\
& \lambda_{1} \leftarrow \lambda_{1} - \sigma \gamma|| \mathcal{C}_{dyn \_ up}||_2 \\
& \lambda_{2} \leftarrow \lambda_{2} - \sigma \gamma|| \mathcal{C}_{dyn \_ low}||_2 \\
& \sigma \leftarrow \rho \sigma , \rho > 1 \\
\end{aligned}
\end{equation}

Minimizing this Augmented Lagrangian function renders the action command safe while minimally deviating from the initial RL policy and obeying the robot dynamic constraints.

\begin{table*}[htbp]
\centering
\vspace{0.15cm}
\begin{tabular}{cccccccccc}
\hline
\multirow{2}{*}{} & \multicolumn{3}{c}{sparse\_4} & \multicolumn{3}{c}{dense\_4} & \multicolumn{3}{c}{empty\_8} \\ \cline{2-10} 
 & \begin{tabular}[c]{@{}c@{}}success \\ rate\end{tabular} & \begin{tabular}[c]{@{}c@{}}collision\\ rate\end{tabular} & \begin{tabular}[c]{@{}c@{}}done\\ time\end{tabular} & \begin{tabular}[c]{@{}c@{}}success \\ rate\end{tabular} & \begin{tabular}[c]{@{}c@{}}collision\\ rate\end{tabular} & \begin{tabular}[c]{@{}c@{}}done\\ time\end{tabular} & \begin{tabular}[c]{@{}c@{}}success \\ rate\end{tabular} & \begin{tabular}[c]{@{}c@{}}collision\\ rate\end{tabular} & \begin{tabular}[c]{@{}c@{}}done\\ time\end{tabular} \\ \hline
FDMCA & \textbf{97.50\%} & 2.50\% & 14.3s & \textbf{97.50\%} & 2.50\% & 12.5s & 91.25\% & 5.00\% & 82.5s \\
MNDS & 37.50\% & 62.50\% & 9.3s & 23.75\% & 76.25\% & 10.8s & 30.63\% & 64.38\% & 79.5s \\
OURS/RL & 96.25\% & 3.75\% & 11.6s & 92.90\% & 6.25\% & 12.9s & 75.63\% & 23.75\% & 74.7s \\
OURS/RL\_CBF\_REFINE & 90.00\% & \textbf{0.00\%} & 27.3s & 80.00\% & \textbf{0.00\%} & 39.5s & 90.00\% & \textbf{0.00\%} & 89.6s \\
OURS/RL\_CBF\_JOINT & \textbf{97.50\%} & \textbf{0.00\%} & 20.7s & 92.50\% & \textbf{0.00\%} & 24.7s & \textbf{96.88\%} & \textbf{0.00\%} & 79.1s \\ \hline
\end{tabular}
\caption{\small{Comparative results. We evaluate all methods in three different scenarios, sparse environment with four agents, dense environment with four agents, and the empty environment with eight agents.
Multiple policy-sharing agents are initialized with random positions in these scenarios. 
Each setup is tested for 20 rounds. 
To clearly prove the training efficiency and safety ensuring ability, all policies used in this comparison are trained for 200 environment episodes.
The results show that our method can achieve very low collision rates with similar or higher level of success rates, using extra time within tolerance.}}
\label{Evaluation}
\vspace{-0.7cm}
\end{table*}

\subsection{Reward Feedback}
To efficiently cooperate RL with CBF, the RL policy should provide better initial value that helps the refining process.
Therefore, we design a new reward $ r_c $ that receives feedback from the CBF module. 
We take the initial value of $\mathcal{L}_{\sigma}(a, \pi_{\phi}(s), \lambda_{ij}, \lambda_{1}, \lambda_{2})$ as $ r_c $, which delivers the deviation degree from the safe constraints and the robot dynamics constraints.
With this reward, the policy network and the CBF module can cooperate efficiently with consensus.

\subsection{Joint Training Algorithm}
We also provide an online training method shown in Algorithm \ref{algorithm_cbf_rl}, where the policy network and the world transition model are trained alternately. 
As for the policy training network, we adopt the framework of TD3 with network structure the same as the one proposed in our previous work\cite{cui2020learning}. 
In the initial stage of training, the prediction accuracy may not be very reliable, so we enable a simple world transition model with quasi-static assumption of the environment, using only affine transformations based on the velocity and action command for prediction.
And as the training goes on, the deep world transition model takes over.

\section{Experiments}
\subsection{Ablation Study}
\textbf{CBF based Optimizer:}
To prove the merits of our CBF-based optimizer, we test the performance of our RL policy with and without the refine module, both in the training and inference procedures. 
The CBF-based optimizer is only used for policy refinement here.
Both of the policies are trained under the same reward function composed of only dense distance-based goal-reaching reward and collision penalty.
To limit the experiment conditions, the CBF-based optimizer provides fixed steps of refinement for the contrast experiment.
Both policies are trained with the same parameters like learning rate and the number of environment episodes. 
They are also trained in the same simulation environments with four randomly initialized robots and other obstacles.
Multiple random seeds are also used for both methods. 

To compare the performance, we use the following metrics:
\begin{itemize}

\item \textbf{\emph{Success Rate:}} The ratio of robots arriving at their own goals without collisions within the time allowed.

\item \textbf{\emph{Collision Rate:}} The ratio of robots colliding with obstacles or other robots before arriving at their goals.

\item \textbf{\emph{Done Time:}} Average time of an episode.

\end{itemize}

As we can see in Fig. \ref{Training_Curves}, the RL policy with CBF refinement significantly outperforms the individual RL policy.
The success rate rises rapidly while keeping a low collision rate.
Proving that the CBF-based optimizer can work as a convincing safe certificate in multi-robot navigation tasks.
Quantitative results in Table \ref{Evaluation} also prove the feasibility and efficiency of the CBF-based optimizer.

As we check on the failure cases, we find that the collision happens when the CBF-based optimizer provides refinement of large or oscillating values, meaning that the policy and the optimizer have got disagreements on the current state disposal.
And due to the limit on the number of refinement steps, the safety violation may not be minimized enough with an unstable poor initial value provided by the separately trained RL policy, resulting in collisions.

\textbf{Joint Training Framework:}
We also evaluate the performance of the joint framework for comparison. 
In the training process, the policy directly uses initial values of the objective function as rewards, instead of the collision penalty used for the other policies.

The results in Fig. \ref{Training_Curves} show that the RL policy trained jointly with CBF can outperform all other methods.
The policy trained in this method can also achieve high success rates with significantly fewer periods.
And because of the consensus between the policy network and the CBF-based optimizer built by the reward feedback, the policy trained in this way can cooperate better with the safety refine module.
Compared with RL policy only with refinement, the descent in the total steps also proves that the CBF feedback for policy training can narrow the gap of cooperation.
\subsection{Comparative Study}
We further provide a thorough evaluation of the performance of our method by comparing it with existing model-free RL approaches. 
In particular, we choose the following two methods as baselines to compare:
\begin{itemize}

\item \textbf{\emph{FDMCA}}\cite{fan2018fully} is a RL-based dynamic collision avoidance algorithm using 3 consecutive 2D laser scans in angle range representation as input.

\item \textbf{\emph{MNDS}}\cite{jin2020mapless} is a costmap-based RL dynamic collision avoidance algorithm using laser scans in disentangled angle range representation as input. 

\end{itemize}

In the training procedure, as we can see in Fig.\ref{Training_Curves}, FDMCA can also reach a high success rate in a quite short time, which may benefit from its straightforward network architecture with fast convergence.
While in the experiments, we can observe that FDMCA may make short-sighted decisions that lead to shaky trajectories.
However, because of the non-intuitive laser representation used in MNDS, it demands much longer training episodes for convergence, which leads to poor performance in limited training episodes.

\begin{figure}[tb]
\centering
\subfigure[sparse\_4]{
\begin{minipage}[t]{0.32\linewidth}
\centering
\includegraphics[width=1in]{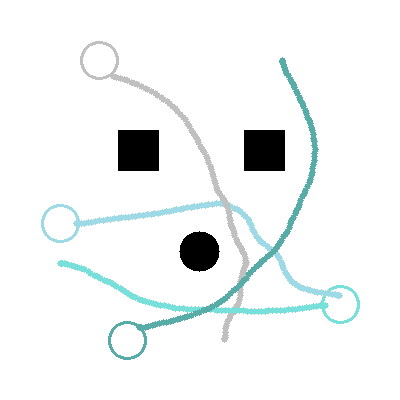}
\vspace{3pt}

\includegraphics[width=1in]{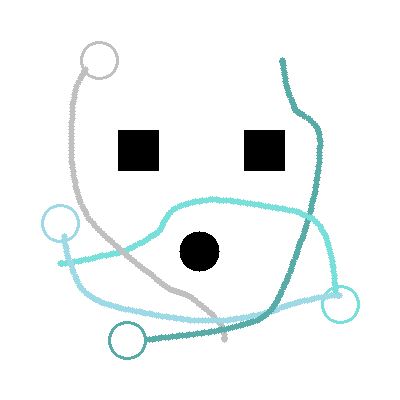}
\vspace{3pt}
\end{minipage}%
}%
\subfigure[dense\_4]{
\begin{minipage}[t]{0.32\linewidth}
\centering
\includegraphics[width=1in]{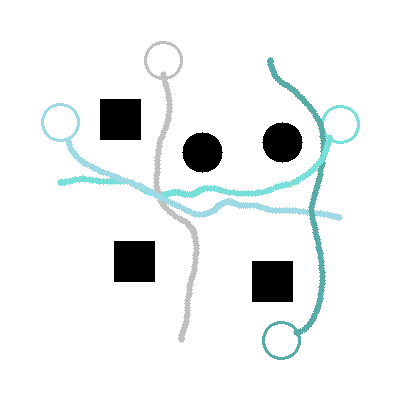}
\vspace{3pt}

\includegraphics[width=1in]{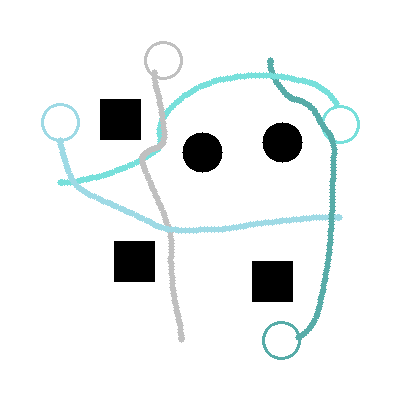}
\vspace{3pt}
\end{minipage}%
}%
\subfigure[empty\_8]{
\begin{minipage}[t]{0.32\linewidth}
\centering
\includegraphics[width=1in]{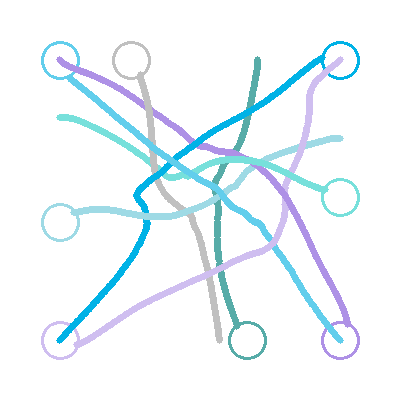}
\vspace{3pt}

\includegraphics[width=1in]{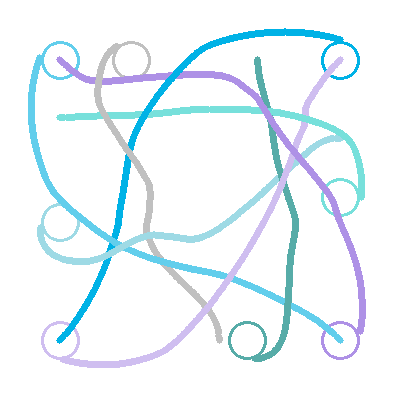}
\vspace{3pt}
\end{minipage}%
}%
\centering
\caption{\small{Trajectory results. Comparison between FDMCA(up) and full version of our method(down) on trajectory level. Circles represent the targets of robots with the same color. Trajectories from our method concern more about safety.}}
\label{traj}
\vspace{-0.2cm}
\end{figure}

The quantitative evaluation results are shown in Table \ref{Evaluation}. It turns out that our methods can ensure the safety of the multi-robot system with both high probability and flexibility using the CBF-based optimizer, under the same training conditions.
The robots can navigate safely using only observation data from the on-board sensors.
In addition, taking advantage of the local observation mechanism, the policy trained in a simple simulation environment can be naturally transferred to unseen, complex, or crowded environments, where robots avoid collisions with similar local readings. In Table \ref{Evaluation}, FDMCA seems to have similar performance with our method, so we do a further comparative experiment between FDMCA and full version of our method on trajectory level. 
Fig. \ref{traj} shows some typical results, in which our method tends to pay more attention to safety. This accords with the results in Table \ref{Evaluation}, which show that our method can avoid crashing cases and is more competitive than others in multi-robot methods.

\subsection{Real World Experiments}
\begin{figure}[tb]
\centering
\subfigure[RL policy]{
\begin{minipage}[t]{0.5\linewidth}
\centering
\includegraphics[scale = 0.25]{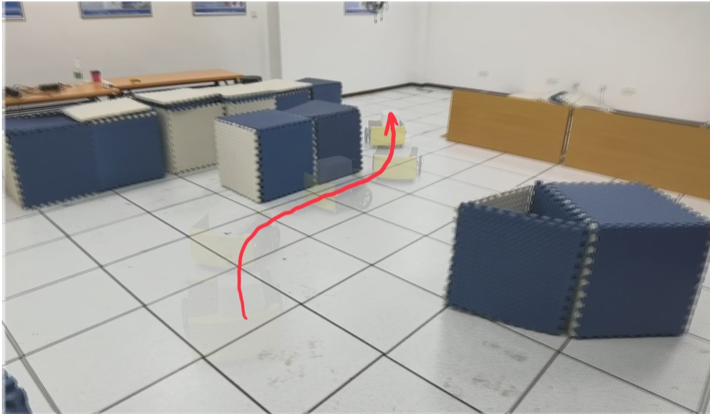}
\vspace{3pt}

\includegraphics[scale = 0.25]{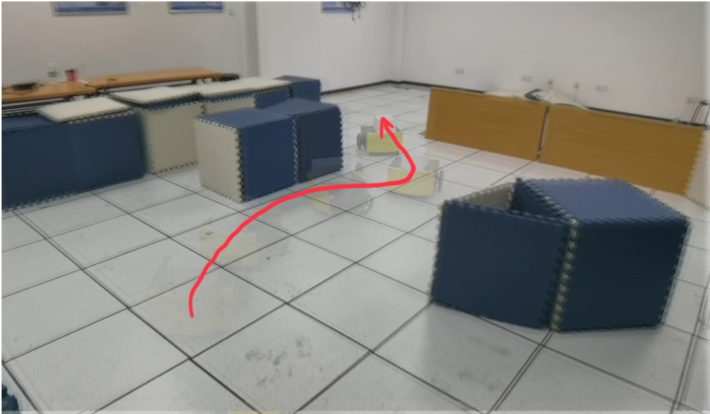}
\vspace{3pt}
\end{minipage}%
}%
\subfigure[RL policy joint with CBF]{
\begin{minipage}[t]{0.5\linewidth}
\centering
\includegraphics[scale = 0.25]{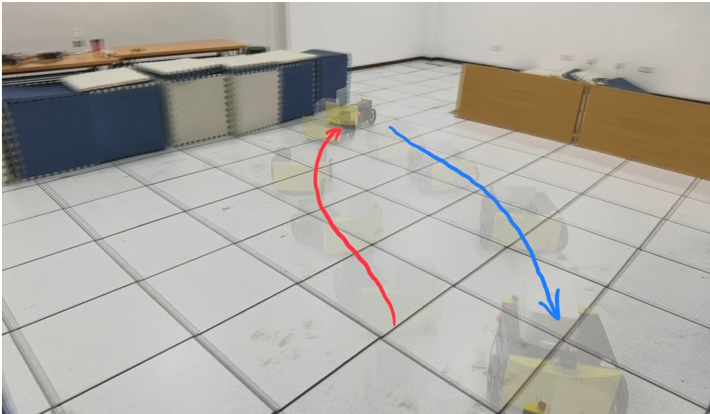}
\vspace{3pt}

\includegraphics[scale = 0.25]{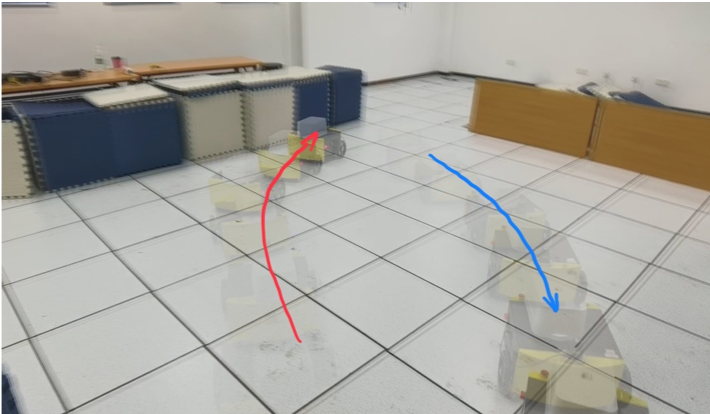}
\vspace{3pt}
\end{minipage}%
}%
\caption{\small{Real robot evaluation. The learned policy is evaluated in multiple real-world scenarios. Comparison results show that the policy learned with the joint framework can complete navigation tasks with stronger safety concerns.}}
\label{real}
\vspace{-0.7cm}
\end{figure}

We evaluate the online performance of the learned policy on real differential robots.
We compare the performance of two methods on the robot, individual RL policy and RL policy joint with the CBF, the results are shown in Fig. \ref{real}. 
Trajectories marked on the images show that the RL policy joint with the CBF achieves safe collision avoidance and target arriving.

\vspace{-0.2cm}
\section{Conclusions}
This paper presents a joint training framework of observation-based CBF with RL for safety certificates in multi-robot navigation. 
With the help of the observation-based CBF and the world transition model, we propose an optimizer that provides both policy refinement for safety certificates and reward feedback for the improvement of policy cooperation. 
Experiments show that the proposed method can ensure safety with both high probability and flexibility in training and inference procedures, and successful deployment in the real world.


\bibliographystyle{ieeetr}

\end{document}